\PassOptionsToPackage{unicode}{hyperref}
\PassOptionsToPackage{hyphens}{url}
\documentclass[
]{article}
\usepackage{amsmath,amssymb}
\usepackage{lmodern}
\usepackage{iftex}
\ifPDFTeX
  \usepackage[T1]{fontenc}
  \usepackage[utf8]{inputenc}
  \usepackage{textcomp} 
\else 
  \usepackage{unicode-math}
  \defaultfontfeatures{Scale=MatchLowercase}
  \defaultfontfeatures[\rmfamily]{Ligatures=TeX,Scale=1}
\fi
\IfFileExists{upquote.sty}{\usepackage{upquote}}{}
\IfFileExists{microtype.sty}{
  \usepackage[]{microtype}
  \UseMicrotypeSet[protrusion]{basicmath} 
}{}
\makeatletter
\@ifundefined{KOMAClassName}{
  \IfFileExists{parskip.sty}{%
    \usepackage{parskip}
  }{
    \setlength{\parindent}{0pt}
    \setlength{\parskip}{6pt plus 2pt minus 1pt}}
}{
  \KOMAoptions{parskip=half}}
\makeatother
\usepackage{xcolor}
\IfFileExists{xurl.sty}{\usepackage{xurl}}{} 
\IfFileExists{bookmark.sty}{\usepackage{bookmark}}{\usepackage{hyperref}}
\hypersetup{
  hidelinks,
  pdfcreator={LaTeX via pandoc}}
\usepackage{longtable,booktabs,array}
\usepackage{calc} 
\usepackage{etoolbox}
\makeatletter
\patchcmd\longtable{\par}{\if@noskipsec\mbox{}\fi\par}{}{}
\makeatother
\IfFileExists{footnotehyper.sty}{\usepackage{footnotehyper}}{\usepackage{footnote}}
\makesavenoteenv{longtable}
\usepackage{graphicx}
\makeatletter
\def\maxwidth{\ifdim\Gin@nat@width>\linewidth\linewidth\else\Gin@nat@width\fi}
\def\maxheight{\ifdim\Gin@nat@height>\textheight\textheight\else\Gin@nat@height\fi}
\makeatother
\setkeys{Gin}{width=\maxwidth,height=\maxheight,keepaspectratio}
\makeatletter
\def\fps@figure{htbp}
\makeatother
\providecommand{\tightlist}{%
  \setlength{\itemsep}{0pt}\setlength{\parskip}{0pt}}

\ifLuaTeX
  \usepackage{selnolig}  
\fi

\title{Quantifying non‑deterministic drift in large language models}
\author{Claire Nicholson\\
HelixScribe.AI, London, United Kingdom\\
\texttt{claire@helixscribe.ai}}
\date{}

\begin{document}

\maketitle

\begin{abstract}

Large language models (LLMs) are employed for tasks from summarisation
to decision support. Identical prompts do not always produce identical
outputs, even when temperature settings are fixed. Here we conduct
repeated‑run experiments and report systematic empirical measurements that quantify baseline behavioural drift, defined as the variability observed when the same prompt is issued multiple times without stabilising instructions. We analyse how these baseline measurements relate to recent drift research. Two publicly accessible models,
\textbf{gpt‑4o‑mini} and \textbf{llama3.1‑8b} were tested on five
categories of prompts under exact repeats, perturbed inputs and reuse
modes at temperatures of~0.0 and~0.7. We measure unique output
fractions, lexical similarity and word count variation, and compare
behaviour across models. Recent literature on concept drift, behavioural
drift and infrastructure‑induced nondeterminism is reviewed to
contextualise the results (Atil et~al.,~2025; Khatchadourian
\&~Franco,~2025). We discuss the limits of lexical metrics and highlight
emerging semantic metrics, and conclude with implications for baseline
measurement and for future mitigation studies (Chen et~al.,~2024;
Maxwell \&~Berenzweig,~2025; Khatchadourian \&~Franco,~2025).

\end{abstract}

\section{Introduction}\label{sec:introduction}

Machine‑learning models trained on historical data often encounter
changes in the relationship between inputs and targets, a phenomenon
known as \emph{concept drift} (Evidently~AI~Blog,~2025). In
classification or regression settings this is usually attributed to
shifts in the data distribution. By contrast, \emph{model drift} refers
to degraded performance or behavioural changes without specifying a
cause (Nicholson,~2025). Large language models introduce an additional
layer: sampling algorithms and serving infrastructure can make a
supposedly deterministic configuration behave unpredictably. In
practice, users expect that temperature~0 and fixed seeds yield
repeatable outputs, yet studies show that is not always the case: top
LLMs did not produce identical completions across ten deterministic runs
and nondeterminism can stem from either the model or the inference stack
(Atil et~al.,~2025; Khatchadourian \&~Franco,~2025). Larger models
tend to exhibit greater variability. For example, a
120‑billion‑parameter model produced identical outputs in only
about~12.5~per~cent of runs. In contrast, 7--8B models were fully
consistent under the same settings, as reported in prior work, and
infrastructure issues such as miscompiled sampling algorithms or
floating‑point non‑associativity can induce random anomalies
(Khatchadourian \&~Franco,~2025). This study therefore asks: \emph{how
much baseline behavioural drift occurs under ordinary, operator‑free
conditions?} \textbf{This work presents a systematic, multi‑mode baseline
quantification of nondeterministic drift across five prompt categories,
two temperatures and two deployment types (API‑served and local).}

\textbf{This manuscript reports new repeated-run experiments and empirical measurements; it is not a review, survey, or position paper.}

Unlike works that benchmark downstream task performance or propose new
control strategies, this paper does not introduce any stabilisation
technique or prompt engineering method. Its sole contribution is the
quantification of baseline variability under an operator‑free regime. By
measuring drift in the absence of interventions we provide a reference
point against which mitigation studies can be compared.

Prompt‑level interventions have been proposed to stabilise LLM
behaviour, such as operator‑level prompts that compress the output
manifold (Nicholson,~2025). Those works focus on reducing variance. Our
focus is complementary: we first establish the magnitude of baseline
behavioural drift under ordinary conditions, a prerequisite for
evaluating control methods. We also draw on recent literature that
distinguishes between measurement‑focused studies (which characterise
drift) and intervention studies (which propose mitigation techniques)
(Khatchadourian \&~Franco,~2025; Atil et~al.,~2025). By clearly
separating measurement from control, researchers avoid masking drift
when evaluating solutions (Khatchadourian \&~Franco,~2025).

While our previous work explored operator‑level prompting as a soft
behavioural control in LLMs (Nicholson,~2025), that study intentionally
compressed the output manifold by injecting structured instructions. The
present paper is conceptually distinct: it does not propose or evaluate
any control mechanism. Instead, it quantifies baseline behavioural drift
in a purely operator‑free regime. This clean separation prevents the
common pitfall of conflating measurement with mitigation and ensures
that subsequent interventions can be assessed against an unbiased
baseline.

\subsection{Novelty relative to prior work}\label{sec:novelty}

Previous studies have investigated nondeterminism in supposedly
deterministic LLM settings by re‑running identical prompts at
temperature~0.0 for a handful of models (Atil et~al.,~2025). Our work differs in three important ways. \textbf{First,}
we examine prompt‑category sensitivity by grouping prompts into five
categories and running repeated experiments for each. \textbf{Second,}
we compare an API‑served model with a local open‑weight model to assess
how deployment type influences baseline drift. \textbf{Third,} we
evaluate three prompting modes (exact repeats, perturbed inputs and
reuse) at two temperatures in a systematic design. Khatchadourian
and~Franco~(2025) conceptualise drift using variance budgets and
cross‑provider validation but do not measure baseline drift across
models or prompt categories. By conducting repeated‑run experiments
across prompt modes, categories and deployment types, our study provides
a systematic multi‑mode baseline quantification of nondeterministic
drift for \textbf{gpt‑4o‑mini} and \textbf{llama3.1‑8b}.

\subsection{Contributions}\label{sec:contributions}

This paper makes the following contributions:

\begin{itemize}
\tightlist
\item
  \textbf{Quantification of baseline drift:} We provide a systematic
  measurement of baseline behavioural drift under operator‑free
  conditions across two publicly accessible models (\textbf{gpt‑4o‑mini}
  and \textbf{llama3.1‑8b}), five prompt categories, three prompting
  modes and two temperatures. These repeated‑run experiments yield lexical
  empirical measurements that serve as a baseline for future
  stabilisation techniques.
\item
  \textbf{Evidence of persistent nondeterminism:} The experiments show
  that nondeterminism persists even at temperature~0.0, with distinct
  variability patterns by model size, deployment and prompting mode.
  This highlights that deterministic settings do not guarantee
  reproducibility and that drift manifests differently across
  categories.
\item
  \textbf{Interpretation via variance budgets:} We discuss how the measured variability can be interpreted using variance-budget concepts from prior work, as a way to connect baseline drift measurements to application-specific tolerance decisions.
\end{itemize}

\section{Definitions and terminology}\label{sec:definitions}

To avoid ambiguity we define key terms used in this paper.

\begin{itemize}
\item
  \textbf{Baseline behavioural drift.} Variability in outputs when
  identical prompts are run repeatedly without any structured operators
  or control. This variation arises from non‑deterministic sampling and
  serving infrastructure.
\item
  \textbf{Operator‑free regime.} A prompting regime with no stabilising
  instructions. All experiments in this study were conducted in this
  regime.
\item
  \textbf{Behavioural drift.} A broader concept referring to progressive
  changes in an LLM's responses or strategies over time. It encompasses
  semantic drift (outputs diverging from the original intent),
  coordination drift in multi‑agent systems and the emergence of new
  unintended strategies (Kim,~2025; Li et~al.,~2025). Behavioural drift
  is often measured over longer horizons or across model versions
  (Kim,~2025).
\item
  \textbf{Variance budget.} A conceptual allowance for how much output
  variability is acceptable. Drawing on reliability engineering, the
  variance budget is analogous to an error budget (Khatchadourian
  \&~Franco,~2025). Khatchadourian and Franco~(2025) propose
  finance‑calibrated thresholds of ±5~per~cent for generated figures;
  exceeding the budget may motivate mitigation. A limited number of
  minor deviations may be tolerated before any corrective action is
  taken (Khatchadourian \&~Franco,~2025).
\item
  \textbf{Behavioural attractor region.} A compact region of output
  space towards which a model converges under certain conditions. Drift
  tails induced by reinforcement‑learning‑from‑human‑feedback (RLHF)
  safety filters can cause subsequent outputs to become overly cautious
  and repetitive (Kim,~2025). Attractor states can also arise in
  multilingual generation when models default to English regardless of
  the input language (Li et~al.,~2025). These attractor regions
  illustrate that drift is not purely random but can reflect stable
  alternative regimes (Kim,~2025; Li et~al.,~2025).
\end{itemize}

Quantifying baseline behavioural drift is a prerequisite for evaluating
any stabilisation technique: without a baseline, one cannot judge
whether an intervention genuinely reduces variability or merely captures
noise.

\section{Methods}\label{sec:methods}

\subsection{Experimental setup}\label{sec:experimental-setup}

This paper is a measurement study. No control or mitigation techniques
were applied; instead, we observe baseline behavioural drift under
ordinary conditions.

We evaluated two publicly accessible LLMs: \textbf{gpt‑4o‑mini}, served
via API, and \textbf{llama3.1‑8b}, an open‑weight model running locally
under three prompting modes (exact repeats, perturbed inputs and reuse
of previous answers) and two temperatures (0.0 and~0.7). These models
were selected deliberately to contrast a smaller API‑served model with a
larger open‑weight model running locally, allowing us to examine how
scale and deployment influence baseline behavioural drift. Five prompt
categories were used: underspecified summaries, underspecified advice,
light constraints, style prompts and hard constraints. Each combination
of model and prompt was run thirty times for the gapfill dataset and
twenty times for the small battery. For perturbed inputs we introduced
synonyms and minor lexical changes; for reuse we fed the previous answer
back into the next prompt. Runs with API errors or refusals were
excluded.

\subsubsection{Implementation details}\label{sec:implementation-details}

For \textbf{llama3.1‑8b} we used the Llama~3~8B base model released in
May~2025. The model was loaded using the vLLM engine and Hugging~Face
Transformers~4.36 on a single NVIDIA~A100 GPU. Inference was performed
with bfloat16 precision and batch size~1. Seeds were fixed across runs.
Decoding parameters were: maximum output length 512 tokens,
top\_p~=~1.0, no top\_k limit, and temperature set to either~0.0 or~0.7
as specified.

For \textbf{gpt‑4o‑mini} we used the \texttt{gpt‑4o‑mini‑20250501}
endpoint of the OpenAI Chat Completions API. Experiments were run
between May and June~2025. Temperature was set to~0.0 or~0.7, with
top\_p~=~1.0 and maximum output length 512 tokens. No streaming was
used. Up to three retries were allowed for transient API errors. Other
infrastructure details (e.g.~batching and sampling algorithm) are opaque
to the user and were left at their defaults.

\subsubsection{Prompt set description}\label{sec:prompt-set-description}

Each prompt category contained 20 distinct prompts in the gapfill
dataset and~5 distinct prompts in the small battery, yielding 100 and~25
prompts respectively. Prompts were manually curated to represent common
interaction types rather than drawn from the model training data. For
example:

\begin{itemize}
\tightlist
\item
  \textbf{Underspecified summaries:}~\emph{``Summarise the plot of a
  well‑known novel in one paragraph.''}
\item
  \textbf{Underspecified advice:}~\emph{``Provide advice for someone
  starting a vegetable garden.''}
\item
  \textbf{Light constraints:}~\emph{``Write a three‑line haiku about
  rain.''}
\item
  \textbf{Style prompts:}~\emph{``Describe your favourite city using a
  formal tone.''}
\item
  \textbf{Hard constraints:}~\emph{``Translate the sentence `The cat
  sits on the mat' into French and ensure the output contains exactly
  two sentences.''}
\end{itemize}

Prompt perturbations were created by substituting synonyms or rephrasing
while preserving meaning. For the reuse mode, the answer from one run
was fed back as input to the next. All prompts and perturbations were
manually generated by the author and did not involve synthetic data.

\subsection{Metrics}\label{sec:metrics}

Three metrics quantify drift in this study:

\begin{enumerate}
\def\labelenumi{\arabic{enumi}.}
\tightlist
\item
  \textbf{Unique output fraction.} The number of distinct answers
  divided by the total runs. It approximates the probability that a
  second call to the model yields a new answer.
\item
  \textbf{Average pairwise Jaccard similarity.} For each prompt we
  tokenised outputs and computed the mean Jaccard similarity between all
  pairs. Jaccard has been used to quantify drift in citation consistency
  and lexical overlap (Chen et~al.,~2024; Atil et~al.,~2025).
  However, lexical metrics operate on surface forms and can overestimate
  drift by flagging harmless paraphrasing (Maxwell \&~Berenzweig,~2025).
  We acknowledge that two outputs can convey the same meaning in
different words; semantic drift metrics are therefore needed to
  complement lexical scores (Maxwell \&~Berenzweig,~2025).
\item
  \textbf{Word count statistics.} Mean word count and its standard
  deviation across runs.
\end{enumerate}

To facilitate replication, all scripts, prompts and run configurations
used in this study will be uploaded to a public Zenodo repository prior
to camera-ready submission or follow-up publications. Recent
measurement‑focused studies propose additional metrics such as Total
  Agreement Rates (TARr and~TARa), which capture exact output agreement
  and answer‑level agreement respectively (Atil et~al.,~2025), and
Purpose Fidelity, which assesses whether the output still serves the
original intent (Maxwell \&~Berenzweig,~2025). These metrics distinguish
between surface‑level and semantic drift (Chen et~al.,~2024; Maxwell
\&~Berenzweig,~2025). Our analysis relies on lexical metrics for
interpretability but recognises their limitations and suggests future
work to incorporate semantic measures. We select lexical metrics
deliberately as a conservative lower bound on behavioural drift:
differences in wording are straightforward to identify, whereas subtle
semantic changes may require domain‑specific evaluation. This baseline
measurement establishes an unambiguous reference; more sensitive
semantic or task‑specific metrics can then be layered on top to detect
deeper forms of drift. Lexical drift provides a conservative indicator
of behavioural drift; the absence of lexical differences does not rule
out semantic divergence. Given compute and time constraints, this
initial study restricts evaluation to lexical metrics and leaves the
incorporation of semantic metrics for future work.

\subsection{Data and code availability}\label{sec:data-and-code-availability}

All scripts, prompts, data and analysis code used in this study will be
uploaded to a public Zenodo repository prior to camera-ready submission
or follow-up publications; an updated version of this paper will include
the DOI.

\section{Results}\label{sec:results}

Under exact repeats at temperature~0.0, \textbf{gpt‑4o‑mini} produced
distinct outputs in roughly 0.24 of runs, whereas \textbf{llama3.1‑8b}
did so in about 0.09 of runs. The average Jaccard similarity across
outputs was 0.89 for \textbf{gpt‑4o‑mini} and 0.97 for
\textbf{llama3.1‑8b}. Figure~1 illustrates that \textbf{gpt‑4o‑mini}
exhibits roughly two and a half times more variability than
\textbf{llama3.1‑8b}.

At temperature~0.0, perturbing inputs increased variability:
\textbf{gpt‑4o‑mini} had a mean unique fraction of 0.57,
\textbf{llama3.1‑8b} 0.27. Reusing previous answers reduced variability
(0.20 and~0.10 respectively). Increasing temperature to~0.7 led to
near‑complete diversity: almost every run yielded a new answer and
lexical similarity fell below~0.52 for all modes. Table~\ref{tab:metrics} summarises the
mean unique fraction and mean lexical similarity across modes and
temperatures. Overall, raising temperature introduces far more drift
than perturbations or reuse. Values reported in Table~\ref{tab:metrics} are means over
all prompts within a category and across all runs. The standard
deviation of unique output fractions across prompts was below~0.05 for
every combination of model, temperature and mode. Confidence intervals
for each cell are provided in the supplementary files in the Zenodo
repository.

\begin{longtable}[]{@{}
  >{\raggedright\arraybackslash}p{(\columnwidth - 8\tabcolsep) * \real{0.1818}}
  >{\raggedright\arraybackslash}p{(\columnwidth - 8\tabcolsep) * \real{0.1818}}
  >{\raggedright\arraybackslash}p{(\columnwidth - 8\tabcolsep) * \real{0.1299}}
  >{\raggedright\arraybackslash}p{(\columnwidth - 8\tabcolsep) * \real{0.2857}}
  >{\raggedright\arraybackslash}p{(\columnwidth - 8\tabcolsep) * \real{0.1948}}@{}}
\caption{Summary of mean unique fraction and mean Jaccard similarity across models, temperatures and prompting modes}\label{tab:metrics}\\
\toprule
\begin{minipage}[b]{\linewidth}\raggedright
Model
\end{minipage} & \begin{minipage}[b]{\linewidth}\raggedright
Temperature
\end{minipage} & \begin{minipage}[b]{\linewidth}\raggedright
Mode
\end{minipage} & \begin{minipage}[b]{\linewidth}\raggedright
Mean unique fraction
\end{minipage} & \begin{minipage}[b]{\linewidth}\raggedright
Mean Jaccard
\end{minipage} \\
\midrule
\endhead
gpt‑4o‑mini & 0.0 & exact & 0.240 & 0.893 \\
gpt‑4o‑mini & 0.0 & perturb & 0.572 & 0.632 \\
gpt‑4o‑mini & 0.0 & reuse & 0.200 & 0.971 \\
gpt‑4o‑mini & 0.7 & exact & 0.987 & 0.518 \\
gpt‑4o‑mini & 0.7 & perturb & 1.000 & 0.440 \\
gpt‑4o‑mini & 0.7 & reuse & 1.000 & 0.706 \\
llama3.1‑8b & 0.0 & exact & 0.093 & 0.966 \\
llama3.1‑8b & 0.0 & perturb & 0.274 & 0.789 \\
llama3.1‑8b & 0.0 & reuse & 0.100 & 0.910 \\
llama3.1‑8b & 0.7 & exact & 0.987 & 0.471 \\
llama3.1‑8b & 0.7 & perturb & 1.000 & 0.403 \\
llama3.1‑8b & 0.7 & reuse & 0.973 & 0.632 \\
\bottomrule
\end{longtable}

\begin{figure}
\centering
\includegraphics[width=5.83333in,height=4.375in]{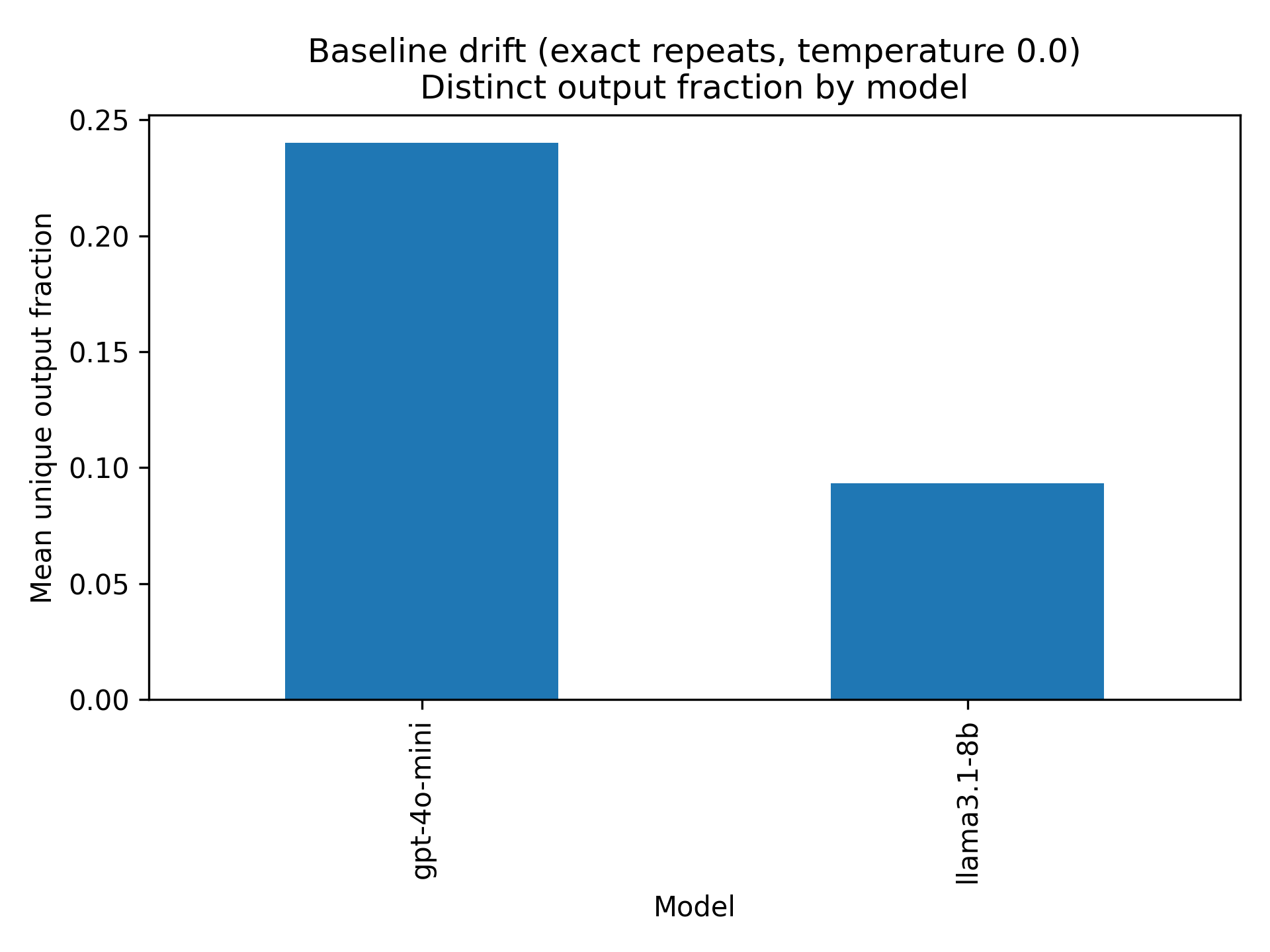}
\caption{Mean unique output fraction for exact repeats at
temperature~0.0}
\end{figure}

\begin{figure}
\centering
\includegraphics[width=5.83333in,height=4.375in]{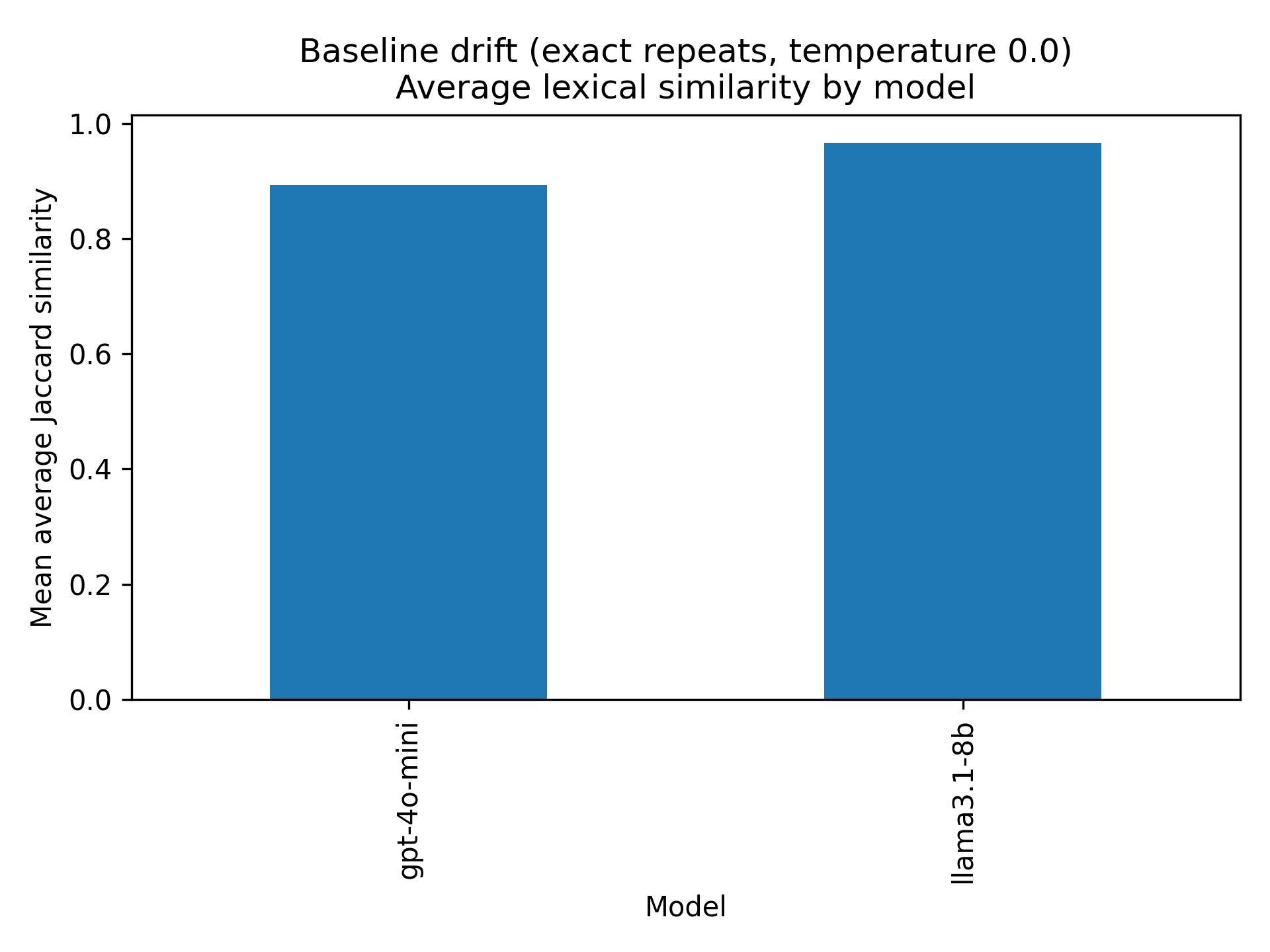}
\caption{Mean average Jaccard similarity for exact repeats at
temperature~0.0}
\end{figure}

\begin{figure}
\centering
\includegraphics[width=5.83333in,height=4.375in]{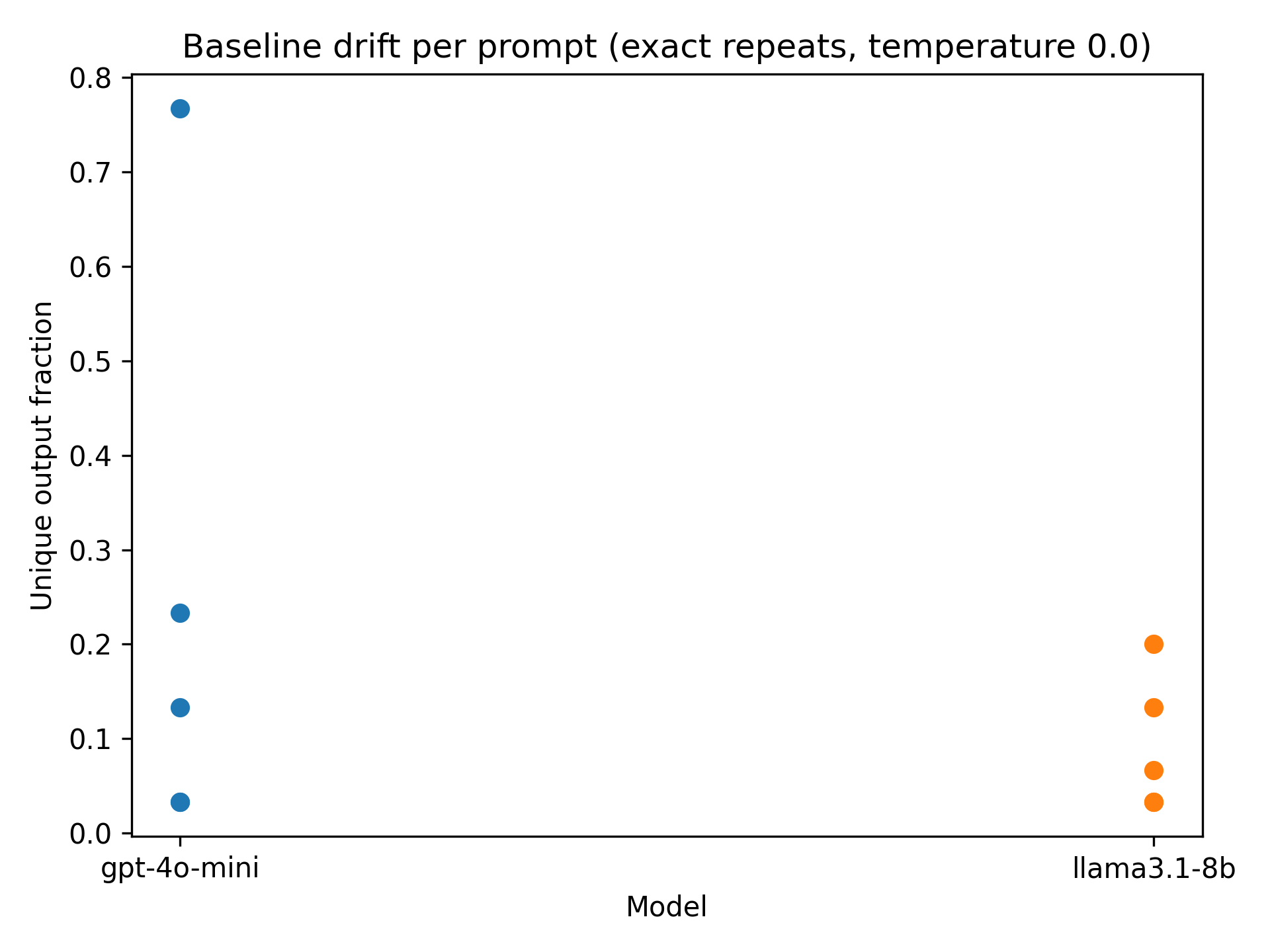}
\caption{Unique output fraction distribution by prompt for exact
repeats at temperature~0.0}
\end{figure}

\section{Discussion}\label{sec:discussion}

\subsection{Drivers of drift}\label{sec:drivers-of-drift}

Our results confirm that baseline behavioural drift exists even at
temperature~0.0: \textbf{gpt‑4o‑mini} produces different outputs roughly
one quarter of the time, while \textbf{llama3.1‑8b} varies in under one
tenth. This observation is consistent with reports from developers and
researchers that a temperature of zero does not guarantee determinism
(Atil et~al.,~2025). Variability increases markedly when inputs are
perturbed or temperature is raised. These findings echo
measurement‑focused studies showing that seemingly deterministic
settings still exhibit accuracy differences across runs and large gaps
between the best and worst run (Atil et~al.,~2025).

The literature situates baseline behavioural drift within a wider
taxonomy. Drift can arise from intrinsic model properties or from
infrastructure. Larger models display more output variance than smaller
ones, and infrastructure issues, including miscompiled sampling
algorithms, batch‑processing effects, hardware bugs and
misconfigurations can induce nondeterminism even when model parameters
and prompts are unchanged (Khatchadourian \&~Franco,~2025). Studies
comparing local and cloud deployments found that local instances can
reproduce deterministic behaviour when seeds, batch sizes and decoding
settings are controlled, whereas API services often introduce extra
variability due to hidden optimisations (Khatchadourian
\&~Franco,~2025). This suggests that some drift observed in practice
reflects the inference stack rather than the model itself.

\subsection{Limitations of lexical metrics}\label{sec:limitations-of-lexical-metrics}

Given that our experiments yielded Jaccard similarities ranging from
0.44 to 0.97 (Table~\ref{tab:metrics}), this section discusses the limitations of using
lexical metrics alone. This study relies on unique output fractions and
Jaccard similarity to quantify drift. Prior work cautions that lexical
metrics are blunt instruments: they are easy to interpret but treat any
difference in wording as evidence of drift, even when meaning is
preserved (Chen et al., 2024; Atil et al., 2025). Paraphrases that
convey the same semantic content will therefore reduce Jaccard
similarity, while more subtle semantic shifts may remain undetected. To
address this limitation, Atil et al.~(2025) propose Total Agreement
Rate metrics (TARr and TARa), which distinguish exact string-level
agreement from answer-level agreement. Maxwell and Berenzweig (2025)
further introduce \emph{Purpose Fidelity}, a measure of whether an
output continues to serve the original intent across generations. There
is growing consensus that surface-level lexical metrics should be
combined with embedding-based or task-specific semantic measures to
distinguish benign variation from meaningful drift. Future work should
incorporate such semantic metrics and may also explore the use of LLMs
themselves as evaluators of output equivalence.

\subsection{Variance budgets and attractor regions}\label{sec:variance-budgets-and-attractor-regions}

Given that our measured unique output fractions ranged from 0.09 to 0.57
depending on temperature and prompting mode (Table~\ref{tab:metrics}), the variance
budget concept frames drift as a resource allocation problem
(Khatchadourian \&~Franco,~2025). Rather than expecting perfect
determinism, one must decide how much variability is acceptable for a
given application. In regulated domains, small thresholds (±5~per~cent)
may be acceptable, and exceeding this budget can trigger corrective
actions (Khatchadourian \&~Franco,~2025). For example, a system that
generates financial figures might set a ±5~per~cent tolerance on word
count or numerical variance. If the observed drift surpasses this
budget, reducing the temperature or applying control prompts has been
reported in prior work to lower variability (Khatchadourian \&~Franco,
2025). Attractor regions explain why outputs sometimes converge to
repetitive patterns: RLHF safety filters can lock a model into overly
cautious responses, and multilingual LLMs may default to English despite
prompts in other languages (Kim,~2025; Li et~al.,~2025). In these cases,
heavy constraints ``use up'' the model's variance budget, concentrating
freedom into a narrow, perhaps undesirable, behaviour (Khatchadourian
\&~Franco,~2025). The appropriate variance budget is application
specific; regulated domains such as finance may tolerate little
variability, whereas creative applications may allow more drift.
Understanding attractors helps interpret drift not merely as random
variation but as movement towards stable alternative regimes (Kim,~2025;
Li et~al.,~2025).

\subsection{Measurement versus control}\label{sec:measurement-versus-control}

Given that unique fractions reach as high as 1.0 at temperature~0.7 and
can be as low as 0.09 at temperature~0.0 (Table~\ref{tab:metrics}), the literature
separates drift measurement from drift mitigation (Khatchadourian
\&~Franco,~2025). Measurement‑focused studies rigorously define and
quantify nondeterminism and often propose metrics such as TARr and TARa
(Atil et~al.,~2025). Intervention‑focused studies introduce
techniques like constrained decoding or drift‑aware training to reduce
variability (Khatchadourian \&~Franco,~2025). Our work is primarily a
measurement study; future research can build upon it to test
interventions. Baseline quantification is a prerequisite for evaluating
stabilisation techniques; without a measured reference, it is impossible
to determine whether an intervention reduces drift or simply obscures
it.

\subsection{Multi-agent and long-horizon drift}\label{sec:multiagent-and-longhorizon-drift}

Although our experiments were single‑turn, the variability we observed
in Section~\ref{sec:results} implies that multi‑agent and long‑horizon interactions may
accumulate drift. As LLMs are increasingly deployed in multi‑agent
systems and over long interactions, drift may accumulate. Rath~(2026)
defines \emph{agent drift} as the progressive degradation of agent
behaviour and introduces the Agent Stability Index; he also proposes
mitigation strategies such as episodic memory resets and adaptive
routing (Rath,~2026). Open questions include how to scale drift metrics
to thousands of turns and whether long‑running systems stabilise or
continue to degrade (Rath,~2026). Proposed techniques like
context‑window management and drift‑aware memory may be necessary to
keep agents on task (Rath,~2026). Future benchmarks should incorporate
multi‑run variability tests and stability scores (Rath,~2026).

\subsection{Limitations}\label{sec:limitations}

This study has several limitations. It focuses on two models,
\textbf{gpt‑4o‑mini} and \textbf{llama3.1‑8b}, and a small set of
prompting categories. The findings may not apply to other architectures,
tasks or larger models. We measure surface‑level lexical variability;
this does not capture semantic equivalence or deeper meaning shifts. The
experiments involve relatively short prompts and single‑turn
interactions, so the results do not address multi‑agent systems or
long‑horizon dialogues. Exploring semantic drift metrics, additional
model scales and deployments, and extended interactions will be
important directions for future work. Closed frontier models served via
proprietary APIs may exhibit different nondeterminism patterns due to
limited reproducibility guarantees and were not evaluated here.

\section{Conclusion}\label{sec:conclusion}

This study quantifies baseline behavioural drift in two LLMs, one API‑served and one open‑weight, and
situates the findings within a broader context. Even with deterministic
settings and identical prompts, \textbf{gpt‑4o‑mini} produces distinct
outputs about one quarter of the time, while \textbf{llama3.1‑8b} varies
in about one tenth. Perturbations and higher temperatures magnify drift,
leading to unique answers in nearly every run. These results indicate
that nondeterminism is an inherent feature of current LLMs and arises
from both model scale and infrastructure. Our findings suggest that
relying on a single run can obscure variability and worst‑case
performance; reporting multiple runs provides a more complete picture.
Lexical metrics served as a starting point in this study; incorporating
semantic measures would provide richer insight into meaning‑preserving
and meaning‑changing divergence. Interpreting drift via a variance
budget can inform decisions about acceptable variability and potential
interventions. Future work could explore mitigation techniques, extend
measurement to multi‑agent and long‑horizon settings and develop
standardised stability benchmarks. Understanding and controlling drift
will be important for improving the reliability of future LLMs. This
work should be read as descriptive rather than prescriptive.

\appendix
\section{Sources of API-induced nondeterminism}\label{app:api-induced-nondeterminism}

Inference through cloud‑hosted APIs may introduce additional sources of
variability beyond model randomness. Hidden optimisations such as
dynamic batching, request queuing and throttling can cause different
contexts to be processed together, altering decoder state and resulting
outputs. Infrastructure‑level factors, including network latency,
hardware heterogeneity and transient memory contents (cache effects),
further affect reproducibility. Some providers may also perform traffic
shaping, contextual reranking or micro‑batching that may change the sampling
distribution across requests. These mechanisms are often opaque, vary
across providers and can shift over time. As a result, API‑induced drift
should be regarded as a systemic property of the inference pipeline
rather than a user‑controlled parameter.

\section{Author contributions and use of AI tools}\label{app:author-ai-tools}

This article was conceived, designed, implemented and written by the
author. Generative AI tools were not used to draft or edit the
manuscript or to decide the experimental design or analysis. Large
language models were used only for routine tasks such as debugging code
and formatting scripts. All experimental decisions, interpretation of
results and writing were done by the author.

\section*{References}

\textbf{Atil, B., Aykent, S., Chittams, A., Fu, L., Passonneau, R.,
Radcliffe, E., Rajan~Rajagopal, G., Sloan, A., Tudrej, T., et~al.}
(2025) \emph{Non‑Determinism of ``Deterministic'' LLM Settings}. arXiv
preprint arXiv:2408.04667.

\textbf{Evidently~AI} (2025) \emph{What is concept drift in ML, and how
to detect and address it}. Evidently~AI~Blog.

\textbf{Statsig~Team} (2025) \emph{Temperature settings: controlling
output randomness}. Statsig Perspectives~Blog.

\textbf{Nicholson, C.} (2025) \emph{Operator‑level prompting as soft
behavioural control in LLMs: evidence from a 7.4× manifold compression}.
Zenodo. https://doi.org/10.5281/zenodo.17716480.

\textbf{Chen, X., et~al.} (2024) \emph{Citation Drift as a
Reproducibility Failure in Scientific LLMs}. In \emph{Proceedings of
NeurIPS~2024}.

\textbf{Khatchadourian, R. \&~Franco, R.} (2025) \emph{LLM Output Drift:
Cross‑Provider Validation \&~Mitigation for Financial Workflows}. In
\emph{Proceedings of ACM~ICAIF~2025}.

\textbf{Clymer, J.} (2025) \emph{Will AI systems drift into
misalignment?} Redwood~Research Blog.

\textbf{Kim, J.} (2025) \emph{Drift Tails in Large Language Models: From
Engineering Awareness to SPC Diagnostics}. Online article.

\textbf{Li, B., et~al.} (2025) \emph{Language Drift in Multilingual
Retrieval‑Augmented Generation}. arXiv preprint arXiv:2511.09984.

\textbf{Maxwell, I. \&~Berenzweig, A.} (2025) \emph{The Half‑Life of
Truth: Semantic Drift vs.~Factual Degradation in Recursive LLM
Generation}. arXiv preprint (forthcoming).

\textbf{Rath, S.} (2026) \emph{Agent Drift: Quantifying Behavioural
Degradation in Multi‑Agent LLM Systems}. arXiv preprint
arXiv:2601.04170.

\textbf{Siciliani, T.} (2025) \emph{Drift Detection in Large Language
Models: A Practical Guide}. Medium~Blog.

\textbf{Zhou, K., et~al.} (2025) \emph{What Prompts Don't Say:
Understanding and Managing Underspecification in LLM~Prompts}. arXiv
preprint arXiv:2505.13360.

\end{document}